\documentclass{bmvc2k}

\usepackage{comment}
\title{CamLessMonoDepth: Monocular Depth Estimation with Unknown Camera Parameters}

\addauthor{Sai Shyam Chanduri}{chanduriss@gmail.com}{1}
\addauthor{Zeeshan Khan Suri}{zshn25@gmail.com}{1}
\addauthor{Igor Vozniak}{Igor.Vozniak@dfki.de}{2}
\addauthor{Christian Müller}{Christian.Mueller@dfki.de}{2}

\addinstitution{
University of Applied Sciences (HTW)\\
Saarbrücken, Germany
}
\addinstitution{
Department of Agents and Simulated Reality (ASR)\\
German Research Center for Artificial\\
Intelligence (DFKI GmbH)\\
Saarbrücken, Germany
}

\runninghead{Chanduri et al.}{CamLessMonoDepth}

\def\eg{\emph{e.g}\bmvaOneDot}

\def\etal{\emph{et al}\bmvaOneDot}

\def\etc{\emph{etc}\bmvaOneDot}

\begin{document}

\maketitle

\begin{abstract}
Perceiving 3D information is of paramount importance in many applications of computer vision. Recent advances in monocular depth estimation have shown that gaining such knowledge from a single camera input is possible by training deep neural networks to predict inverse depth and pose, without the necessity of ground truth data. The majority of such approaches, however, require camera parameters to be fed explicitly during training. As a result, image sequences from wild cannot be used during training. While there exist methods which also predict camera intrinsics, their performance is not on par with novel methods taking camera parameters as input. In this work, we propose a method for implicit estimation of pinhole camera intrinsics along with depth and pose, by learning from monocular image sequences alone. In addition, by utilizing efficient sub-pixel convolutions, we show that high fidelity depth estimates can be obtained. We also embed pixel-wise uncertainty estimation into the framework, to emphasize the possible applicability of this work in practical domain. Finally, we demonstrate the possibility of accurate prediction of depth information without prior knowledge of camera intrinsics, while outperforming the existing state-of-the-art approaches on KITTI benchmark.

\end{abstract}
\vspace{-10pt}
\section{Introduction}
\label{sec:intro}

Perceiving accurate depth is a prerequisite for many application domains like robotics and autonomous driving. Traditionally, such systems rely on information fused from depth sensors such as LiDAR, because of their accuracy and robustness. However, such sensors have limited applicability in extreme weather conditions (fog or heavy rain), limited range, higher cost, and complexity~\cite{dudek2010computational, johnston2020self}. Estimating depth from RGB images can partially mitigate some of these limitations~\cite{godard2019digging}. While supervised learning methods~\cite{eigen, eigen2015predicting, wang2015designing, fu2018deep, guo2018learning} have shown to estimate depth without sensors during inference, they still need ground truth supervision to complete the training. However, acquiring such a large ground truth dataset is a formidable challenge, moreover, expensive and time-consuming. The utilization of synthetic datasets with "free" ground truth data is a possible solution. However, such dataset are lacking representability compared to natural dataset, owing the difficulty in generating photo-realistic synthetic images~\cite{lopez2020desc}. Recent advancements in deep learning helped in alleviating these issues by estimating depth from unlabelled image sequences alone, leading to self-supervised approaches like~\cite{zhou2017unsupervised, godard2017unsupervised, yin2018geonet, godard2019digging}. 

Monocular self-supervised approaches offer a more attractive solution than stereo approaches because of the widespread availability of image sequences available for training. In addition, these monocular approaches require no synchronization of cameras. However, they come with their own set of challenges. They operate under a static scene assumption, meaning, the camera is moving, and the scene is static and violation of such assumption predicts 'holes' in depth maps~\cite{godard2019digging}. They also suffer from issues including scaling to metric-depth when ground truth data is not available, brightness changes because of non-Lambertian and reflective surfaces and, occlusions. Over time, many methods with complex network architectures~\cite{zhou2017unsupervised, guizilini20203d}, engineered loss functions~\cite{godard2019digging,shu2020feature}, masking moving objects~\cite{zhou2017unsupervised,luo2018single, godard2019digging}, using ground truth supervision~\cite{guizilini20203d} or post-processing~\cite{mccraith2020calibrating,xue2020toward} for addressing scale ambiguity, etc., have tried to address such challenges. Yet, current monocular approaches significantly lag behind their counterpart supervised approaches like~\cite{fu2018deep,alhashim2018high, andraghetti2019enhancing}.

Another main limitation in many of the earlier approaches~\cite{zhou2017unsupervised, godard2019digging, yin2018geonet, casser2019depth} is the necessity of precise camera calibration parameters as input, for training accurate depth estimation models. This eliminates the plausible usage of potential data from the wild for training. Previous works~\cite{gordon2019depth, chen2019self, tosi2020distilled} have addressed such issues in the past, however, are not on par with the other approaches which use ground-truth calibrated data. In our work, we primarily address this issue by removing the necessity of pre-calibrated data and also focus on refining monocular depth estimation accuracy. The contributions of this work are fourfold: \textbf{First}, we learn depth from monocular image sequences even when camera intrinsics are unknown, in a self-supervised fashion, motivated from~\cite{gordon2019depth}. We combine it with minimum re-projection loss and auto masking, as proposed by Godard \etal~\cite{godard2019digging}, in order to deal with occlusions and static pixels. \textbf{Second}, we reiterate the importance of using Efficient Sub-Pixel Convolution Networks (ESPCN) (adapted from~\cite{shi2016real, aitken2017checkerboard}) for upsampling purposes to obtain sharper depth estimates. Such a method can leverage super-resolution, that can be more accurate in comparison with interpolation approaches~\cite{shi2016real}. \textbf{Third}, we extend our approach to estimate heteroscedastic pixel-wise uncertainty for depth map, capturing by brightness changes due to specular regions or sudden illumination changes, etc. Such information can be leveraged by the agents to take optimal decisions when they are under-confident in predicting depth values in unknown environments, which can avoid fatalities. For example, in an autonomous driving scenario, uncertainty estimation could be beneficial in handling erroneous estimations from the system to prevent accidents~\cite{poggi2020uncertainty}. \textbf{Finally}, with our exhaustive experiments, we demonstrate that our models, to our best knowledge, outperform the state-of-the-art in monocular depth estimation and closes-in the gap with full-supervised methods on the standard KITTI benchmark~\cite{geiger2013vision} even without the need of camera intrinsics as input.

%-----------------------------------------------------------------------
\vspace{-10pt}
\section{Related Work}
\label{sec:rel_work}
\textbf{Monocular Depth Estimation}. Monocular depth estimation is an ill-posed problem, as multiple plausible depths could correspond to the same pixel on the image plane. Recent works showed the possibility of accurate depth estimation by analysing patterns in appearances using image sequences alone. For instance, Eigen \etal~\cite{eigen} proposed the first deep learning method using multiscale CNNs which regressed depth output by taking only a single image as input. Many such subsequent supervised approaches like~\cite{wang2015designing, fu2018deep, guo2018learning} were later proposed, which further extended this formulation. These approaches demand huge amounts of training data, where obtaining ground truth data is both expensive and time-consuming. To mitigate this problem,~\cite{mayer2018makes} was proposed which makes the use of synthetic data for generating ground truth. However, it lacks representability compared to the natural training data. To tackle this constraint,~\cite{garg2016unsupervised, godard2017unsupervised} were proposed by utilizing the self-supervised strategy that involved learning depth using stereo images for monocular depth inference. In a nutshell, the right images were warped onto the left images using a differentiable sampler as in~\cite{jaderberg2015spatial} which enables learning the depth in an end-to-end manner. Later, Zhou \etal~\cite{zhou2017unsupervised} extended this idea by proposing a strategy to learn the depth along with the pose, to handle training completely with monocular settings only. Furthermore, Godard \etal~\cite{godard2019digging} proposed another set of extensions, where the multiscale approach and per-pixel minimum reprojection loss were adopted for the better handling of occlusions. Other self-supervised monocular depth estimation methods like~\cite{vijayanarasimhan2017sfm, yin2018geonet, ranjan2019competitive, casser2019depth, pillai2019superdepth, guizilini20203d} were proposed over the time, which included more robust architectures, additional loss terms and constraints.

\textbf{Learning from Videos in Wild}. Most of the self-supervised learning works~\cite{zhou2017unsupervised, godard2019digging, yin2018geonet, casser2019depth} require camera intrinsics to learn depth, making it difficult to train on multiple datasets at a time. While Ranftl \etal~\cite{midas} proposed a method to train on multiple datasets at once, their approach to learn depth is supervised, requiring ground-truth depth maps. The lack of camera parameters in generic videos, for instance, a YouTube video, where image sequences are captured from unknown camera setups, limits the usage of such data for training. Traditionally, camera parameters are estimated using techniques which involve calibration targets~\cite{mei2007single, shah1994simple, ying2008identical}, geometric structures~\cite{barreto2005geometric, zhang2015line} or separate neural networks~\cite{bogdan2018deepcalib, Facil_2019_CVPR} and are fed as input to the depth estimation models. The main disadvantages of using such approaches include either the necessity of additional data (for calibration), geometric assumptions, or additional complexity and training time. A recent work, proposed by Gordon \etal~\cite{gordon2019depth} eliminates this need of pre-calibrated camera parameters as input for depth estimation, where camera parameters along with depth, pose and object motion are simultaneously learnt in a fully self-supervised manner. As camera parameters are learnt in a fully self-supervised manner, even image sequences from wild can be used, where the network generalizes better. Later, other works, applicable to image sequences from wild, such as in~\cite{chen2019self} and~\cite{tosi2020distilled} were proposed which followed an approach similar to~\cite{gordon2019depth} in camera parameters estimation. Our approach is similar to these, but does not need object motion mask as input~\cite{gordon2019depth, tosi2020distilled} or online refinement~\cite{chen2019self}, to output accurate depth maps, while assuming a pinhole camera model with minimum or no distortion.

\textbf{Depth Super-resolution}. The problem of image enhancement has been a primary challenge in the fraternity of computer vision. Approaches like~\cite{dong2014learning,simonyan2014very,kim2016deeply, ledig2017photo} have shown to solve this problem while improving upon traditional methods like interpolation. However, such methods demand huge training times and are infeasible when used in combination with applications like depth estimation. For faster and accurate results, Shi \etal~\cite{shi2016real} proposed ESPCN and later~\cite{pillai2019superdepth} has extended it to depth estimation, but requires camera parameters as input. In our work, we exploit ESPCN along with initialization technique from~\cite{aitken2017checkerboard} to remove checkerboard artifacts. Consequently, it improves depth estimation accuracy and also, enables faster training and inference compared to other interpolation approaches~\cite{shi2016real}. 

\textbf{Uncertainty}. The two types of uncertainties in Bayesian world, namely epistemic and aleatoric, are discussed in recent works~\cite{kendall2017uncertainties, kendall2018multi}. Later Klodt \etal~\cite{klodt2018supervising} proposed an approach to model aleatoric uncertainty which also served for increasing depth estimation accuracy. In this paper, the authors have used a Bayesian framework to model photometric uncertainty as predictive variance. Another approach~\cite{johnston2020self} made use of discrete disparity volume module and to model depth uncertainties. Later, Poggi \etal~\cite{poggi2020uncertainty} explored various approaches including both empirical and predictive techniques for depth uncertainty estimation in self-supervised methods. Inspired from~\cite{kendall2018multi}, we propose an approach to model heteroscedastic aleatoric uncertainty as photometric variance by considering noise inherent in the input data.

%The static scene assumption of self-supervised monocular depth estimation methods is violated by moving objects in the scene. To deal with such uncertainties, we propose to learn as soft uncertainty mask within our training pipeline.

%-------------------------------------------------------------------------
\vspace{-3mm}
\section{Methodology}
\label{sec:method}

\begin{figure*}
\begin{center}
\includegraphics[width=\textwidth]{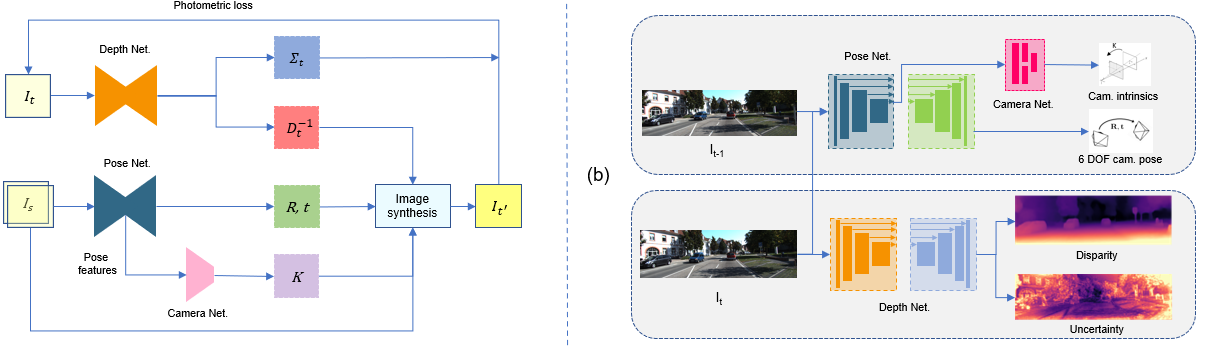}
\end{center}
\vspace{-10pt}
  \caption{(a) shows the overall pipeline used in our approach. (b) depicts the inference model where the depth model can be used to predict inverse-depth and uncertainty from a single image. The pose and camera models can be used to predict camera pose and intrinsics respectively from two temporal frames.}
\label{fig method}
\vspace{-5mm}
\end{figure*}

Self-supervised depth estimation approaches exempt the necessity of hard ground truth by generating a supervision signal using a moving camera setup and by measuring the overlap between temporal image sequences. This Structure-From-Motion (SFM) problem is modelled by performing a novel image synthesis of the target frame from image source point of view. However, this objective is ill-posed since knowledge of accurate pose and camera intrinsics can reconstruct this novel view even with incorrect depth estimations~\cite{godard2019digging}. To address such ambiguity, appearance matching and smoothness losses were proposed. However, it demands the knowledge of pre-calibrated camera intrinsics for novel view synthesis.

Motivated from earlier approaches~\cite{godard2019digging,gordon2019depth}, we propose a framework to simultaneously learn depth, camera pose, uncertainty and camera intrinsics from input video sequences, as seen in \autoref{fig method}(a). The depth network predicts a disparity map (inverse depth) $D_t^{-1}$ of the target image $I_{t}$ and it's corresponding uncertainty map $\Sigma_t$. Simultaneously, the pose network takes any two consecutive frames $I_s$ (either $I_{t+1}$ or $I_{t-1}$) and $I_t$, outputs a 6 Degrees-of-Freedom (DOF) rigid transformation $T_{t\rightarrow s} \in \text{SE(3)}$ from the target image plane to the source image plane, containing rotational $R \in \text{SO(3)}$ and transitional $t \in \mathbb{R}^3$ information.
% Here, depth network takes in a single target image $I_{t}$ and predicts uncertainty $\Sigma_{t}$ and disparity or inverse-depth as output. In order to prevent the gradient locality of the bilinear sampler used for upsampling depth images for computing photometric and to prevent the loss objective in local minimum, we adapt multi-scale decoder from~\cite{godard2019digging} which outputs both, disparity and uncertainty at multiple scales (refer \autoref{fig method}(c)). The disparity obtained is later converted to depth using $D_{t} = 1/(a*\sigma_{t}+b)$ where $\sigma_{t}$ represents disparity map with a and b are chosen to constrain the depth values between 0.1 and 100. 
The camera network takes the pose encoder features as input and outputs the camera intrinsics matrix $K$, containing principal offset and focal length information. After obtaining these predictions, we reconstruct $I_t$ from the source image's point of view (POV). This novel view synthesis involves lifting the target image into 3D using the predicted depth map $D_t$ and the intrinsics matrix's inverse $K^{-1}$. Then, using the predicted transformation matrix $T_{t\rightarrow s}$, and K, we project the 3D scene onto the source's POV, thus warping the source image into the target image~\cite{zhou2017unsupervised, godard2019digging}. We call this warped target image $I_{t'}$. Such a transformation from the target image homogeneous coordinates $p_t$ to the source ones $p_s \in \mathbb{R}^3$, is summarized by the relation $p_s \mbeq \langle KRK^{-1}D_{t}(p_t)p_t + Kt\rangle$~\cite{zhou2017unsupervised}, where $\langle \rangle$ denotes the sampling operator, forms the backbone of this approach.

Relying on a static scene assumption, the warped target image $I_{t'}$ is compared against the original target image $I_t$, using a photometric error metric $\mathcal{L}_{p}(I_{t}, I_{t'})$, which is minimized. Following previous approaches~\cite{zhou2017unsupervised, godard2019digging}, we use a convex combination of L1 loss and a structural similarity(SSIM) loss~\cite{wang2004image} for calculating the photometric error, with $\alpha=0.85$. In order to handle occlusions, we follow Godard \etal's~\cite{godard2019digging} use of per-pixel minimum reprojection which compares views generated from multiple source frames to the target frame and considers the minimum of the per-pixel photometric error, as in \autoref{eqn photo}. This computation prevents high error values when correspondences are good~\cite{godard2019digging}.

%\vspace{-3mm}
\begin{equation}
\mathcal{L}_{p}(I_{t}, I_{s}) = \underset{I_{s}}{min}\;\mathcal{L}_{p}(I_{t}, I_{t'}) = \underset{I_{s}}{min}\;\left (\alpha \; \frac{1-SSIM(I_{t}, I_{t'})}{2} + (1- \alpha)\left \| I_{t}- I_{t'} \right \|_{1} \right)
\label{eqn photo}
\end{equation}

We learn a heteroscedastic Bayesian uncertainty map $\Sigma_t$, for input $I_t$, $\sigma_{ij} \in [0,1] \ \forall \ \sigma_{ij} \in \Sigma_t$, capturing the photometric variance caused by specular objects or by sudden illumination change, etc., following from Kendall \etal's~\cite{kendall2018multi} residual weighting, as shown in \autoref{eqn uncert}. Because of such modelling, the pixels corresponding to such brightness changes are down weighted~\cite{yang2020d3vo} and as a result, the uncertainty values are higher at regions with high photometric reprojection error.

\begin{equation}
\mathcal{L}_{p}(I_{t}, I_{s})^* = \frac {\underset{I_{s}}{min} \; \mathcal{L}_{p}(I_{t}, I_{t'})}{2\Sigma_t^2} + \frac{1}{2} \; \log {\Sigma_t}+1.5
\label{eqn uncert}
\end{equation}

Real world scenes do not always follow a static scene assumption, which is necessary for self- supervised monocular depth estimation. The performance of these methods suffers greatly in presence of static camera or object motion in same direction as camera which can manifest infinite depth results~\cite{godard2019digging}. To handle such scenarios where the relative motion between camera and objects is zero, we make use of binary mask $\mu \in$ \{0,1\} from~\cite{godard2019digging} to filter out static pixels shown in \autoref{eqn bin_mask}. Here [ ] represents the Iverson bracket. Such static pixels are removed by identifying pixels with higher warped reprojection loss between the warped source frame $I_{t^{'}}$ and target frame $I_t$ than the unwarped reprojection loss between the source and target frames.
\begin{equation}
\mu = \left [ \underset{I_{s}}{min} \; \mathcal{L}_{p}(I_{t}, I_{s}) \; > \; \underset{I_{s}}{min} \; \mathcal{L}_{p}(I_{t}, I_{t'})\right]
\label{eqn bin_mask}
\end{equation}

In order to encourage the smoothness of disparity estimations, especially in textureless and low-image gradient areas, and for regularizing inverse depth to prevent divergent values, we add a combination of first and second order smoothness losses $\mathcal{L}_{s}$. \autoref{eqn disp_loss} represents such a disparity smoothness term, weighted by the image's gradients, in order to preserve edges. The first-order gradient term (from) $\nabla_{1}\equiv \partial_{x}+ \partial_{y}$, is an L1 penalty on disparity gradients used to account for the depth discontinuities which often occur at strong image gradients~\cite{heise2013pm}. Adding a second order gradient $\nabla_{2}\equiv \partial_{xx}+ \partial_{xy}+ \partial_{yx} + \partial_{yy}$, in addition, encourages better smoothness of these gradients with larger convergence radii, leading to better optimization~\cite{shu2020feature}. To discourage shrinking of the estimated depth~\cite{godard2019digging}, mean normalized inverse depth $\hat{D_{t}^*}=D_{t}^{-1}/\ \overline{D_{t}}$ is considered, where $D_{t}^{-1}$ represents inverse depth or disparity and $\overline{D_{t}}$ represents the mean disparity.
%Because of downsampling at multiple stages in the depth decoder, this loss is decayed by a factor of 2 for each pyramid level.

\begin{equation}
\mathcal{L}_{s} = \left | \nabla_{1}\hat{D_{t}^*} \right | e^{-0.5\left | \nabla_{1}I_{t} \right |} + \left | \nabla_{2}\hat{D_{t}^*} \right | e^{-0.5\left | \nabla_{2}I_{t} \right |}
\label{eqn disp_loss}
\end{equation}

\textbf{Final Training Loss}. The final SFM objective now, is the combination of the net per-pixel minimum photometric loss with uncertainty (\autoref{eqn uncert}) and the disparity smoothness loss (\autoref{eqn disp_loss}) given by $\mathcal{L}_{t} = \mu \mathcal{L}_{p}^* + \lambda \mathcal{L}_{s}$. Here, $\lambda$ is the smoothness regularizer, to be chosen as a hyperparameter. This total loss $\mathcal{L}_{t}$ is averaged over the total number of scales used in depth decoder, which is $4$ in our work.

\textbf{Network details}. The depth and pose networks follows U-net type encoder-decoder architecture with skip connections in between the encoder and the decoder which facilitates better learning of deep abstract features along with spatial information. We use ResNext-50~\cite{xie2017aggregated} \textbf{depth encoder} with a cardinality of 32 and base width of 4 and ResNet-50 \textbf{pose encoder}, unless stated otherwise. Following earlier works like~\cite{guo2018learning,godard2019digging, kuznietsov2017semi, johnston2020self} in depth estimation, we also employ pre-trained ImageNet~\cite{russakovsky2015imagenet} weights for initialization for the encoder. However, we use weights pre-trained in a semi-weakly supervised fashion as proposed in~\cite{yalniz2019billion} for both depth and pose networks unlike earlier works~\cite{godard2017unsupervised,godard2019digging,guizilini20203d}, which has improved depth estimation accuracy for us. \textbf{Depth decoder} uses a multiscale architecture to output inverse-depth and uncertainty at $4$ scales similar to~\cite{godard2019digging} with two primary modifications: (i) nearest-neighbour interpolation replaced with ESPCN for upsampling following ideas from~\cite{shi2016real} and~\cite{aitken2017checkerboard} and, (ii) modification of final layer in the decoder to output depth uncertainty along with disparity. \textbf{Pose decoder} follows architecture from~\cite{godard2019digging}. \textbf{The camera network} is inspired from~\cite{gordon2019depth} but the architecture is slightly different. More network details are mentioned in supplementary section. The features obtained from the pose encoder are passed to this camera network, followed by squeeze operation to reduce the number of channels to 256. Two independent 3x3 convolution layers stem from this squeeze layer to estimate the normalized focal lengths $f_x,f_y$ and principal offsets $c_x,cy$ in horizontal and vertical axes, normalized by the input image's width and height respectively. These are concatenated to output the camera intrinsics matrix $K = \Big(\begin{smallmatrix}
  f_x & 0 & c_x\\
  0 & f_y & c_y\\
  0 & 0 & 1
\end{smallmatrix}\Big)$. The softplus activation function $f\left(x\right) = \log\left(1+\exp\left(x\right)\right)$ is used to avert negative values for focal lengths. %As these estimations are per-pixel estimations, they are scaled to the input resolution each time.
% Each of these parameters are estimated as per-pixel values, hence are needed to be scaled to the input resolutions accordingly each time. Further network details are discussed in detail in supplementary material.

\textbf{Efficient Sub-pixel convolutions}. Since we use an encoder-decoder architecture for depth, it is necessary to upsample the outputs in decoder layers. Earlier approaches like~\cite{godard2019digging, zhou2017unsupervised, wang2018learning, yin2018geonet} rely on a nearest-neighbour interpolation for faster inference. Such techniques could compromise the output at object boundaries, as they combine the values from foreground and background. Moreover, the filters are not learnable, thus, limiting the upsampling effect. Transpose convolutions, on the contrary, are learnable, but suffer from checkerboard artifacts. To compensate for these artifacts, usually they are initialized with interpolation outputs, which come at the cost of training time. Hence, we make use of ESPCN~\cite{shi2016real} to perform convolutional learning at low resolution. We then perform a pixel shuffle operation to upsample only on the final step. To alleviate the checkerboard artifacts caused by random initialization of these filters, we make use of Initialization to Convolution Nearest-neighbour Resize (ICNR) from~\cite{aitken2017checkerboard} which provides an initialization similar to nearest-neighbour. Such operation results in faster convergence and leads to a better minimum compared to transpose convolutions and other interpolation approaches~\cite{shi2016real}.
%-----------------------------------------------------------------------

\vspace{-13pt}
\section{Experiments}
\vspace{-5pt}
\label{sec:expts}
%\subsection{Setup}
%\label{subsec:setup}
%\vspace{-5mm}
\textbf{Setup.} Our PyTorch~\cite{NEURIPS2019_9015} models were trained on 4 Nvidia GeForce RTX 2080 Ti GPUs, in a distributed setting, using the Adam optimizer~\cite{kingma2014adam}, with a learning rate of $10^{-4}$ for the first $15$ epochs and $10^{-5}$ for the last $10$. The batch size is set to 12 and the default resolution is $640$x$192$ (width x height), unless stated otherwise. The models are trained with data augmentation with a $50 \%$ chance of horizontal flips, random brightness ($\pm 0.2$), random contrast ($\pm 0.2$), hue jitters ($\pm 0.1$) and, saturation ($\pm 0.2$) variations. The input trio of frames at times $t-1$, $t$, and $t+1$, are all applied with the same augmentation settings, but these augmentations are not used for computing the photometric loss. The max. and min. depths are set to 0.1 and 100 respectively. The remaining training parameters are set as in~\cite{godard2019digging}.

\vspace{-10pt}
\subsection{KITTI Dataset}
\label{subsec:kitti}
\begin{figure*}[ht]
\begin{center}
\includegraphics[width=\textwidth]{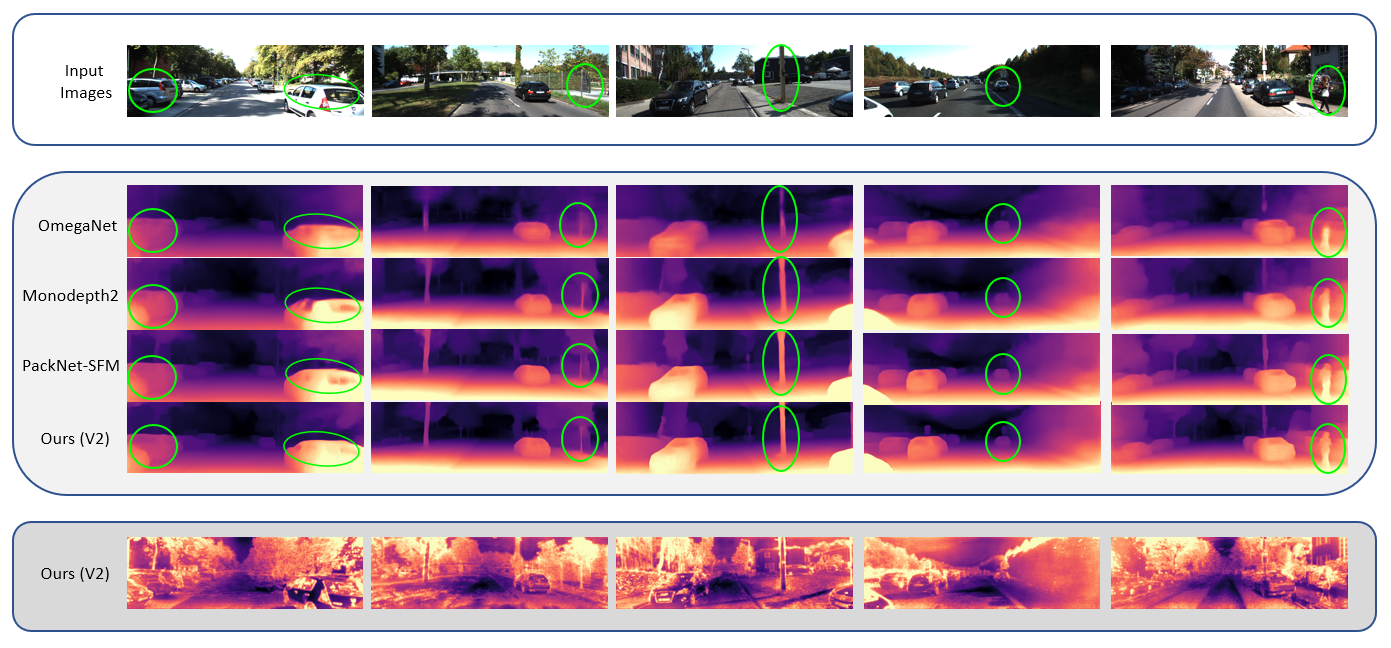}
\end{center}
\vspace{-15pt}
  \caption{The input images, their corresponding disparity maps, and uncertainty maps (bottom) are shown with blocks. Our depth results are compared with MonoDepth2~\cite{godard2019digging}), OmegaNet~\cite{tosi2020distilled} and PackNet-SFM~\cite{guizilini20203d}. For disparity maps, brighter pixels(orange) indicate nearer regions while darker indicate for farther. It can be observed that our approach shows high-quality depth maps especially at object boundaries. For uncertainty maps, brighter regions(orange) indicate pixels with higher uncertainty. Such uncertainty is more pronounced along object boundaries, when further from the centre of camera, also where brightness change occur, \eg at reflective surfaces (a).}
\label{fig qual_res}
\end{figure*}
\vspace{-10pt}

\begin{table}[!htbp]
\begin{center}
\begin{adjustbox}{max width=\textwidth}
\begin{tabular}{cccccccccc}
\hline
\toprule
& &\multicolumn{4}{c}{Lower is better} &\multicolumn{3}{c}{Higher is better} \\
\cmidrule(r){3-6}\cmidrule(l){7-9}
Method   & Intr  &Abs Rel   &Sq Rel  &RMSE  &RMSE log  &$\delta$ $<$ 1.25  &  $\delta$ $<$ $1.25^{2}$  &$\delta$ $<$ $1.25^{3}$ \\
\bottomrule
\hline
SFMlearner~\cite{zhou2017unsupervised} & \xmark	&0.208	&1.768	&6.958	&0.283	&0.678	&0.885	&0.957 \\
MonoDepth2~\cite{godard2019digging}	& \xmark	&0.115	&0.903	&4.863 	&0.193	&0.877	&0.959	&0.981 \\
PackNet-SFM~\cite{guizilini20203d} & \xmark	&0.111	&0.785	&4.601	&0.189	&0.878	&0.960	&0.982 \\
HR-Depth~\cite{lyu2020hr}  & \xmark &0.109	&0.792	&4.632	&0.185	&0.884	&0.962	&0.983 \\
Johnston \etal~\cite{johnston2020self} & \xmark &0.106	&0.861	&4.699	&0.185	&0.889	&0.962	&0.982 \\
\midrule
GLNet (-ref)~\cite{chen2019self} & \cmark &0.135 &1.070 &5.230 &0.210 &0.841 &0.948 &0.980 \\
Gordon \etal~\cite{gordon2019depth}	& \cmark	&0.128	&0.959	&5.230	&0.212	&0.845 	&0.947	&0.976 \\
OmegaNet~\cite{tosi2020distilled} & \cmark	&0.120	&0.792	&4.750	&0.191	&0.856	&0.958	&\textbf{0.984} \\ % (L) GLNet(-ref.) X 0.135 1.070 5.230 0.210 0.841 0.948 0.980
\midrule
\textbf{Ours without uncert. (V1)}       & \cmark      & \textbf{0.105}   & 0.768  & 4.539  & \textbf{0.182}   & 0.890        & \textbf{0.964}           & 0.983          \\
\textbf{Ours with uncert. (V2)}    & \cmark          & 0.106   & \textbf{0.750}   & \textbf{4.482}  & \textbf{0.182}    & \textbf{0.891}        & \textbf{0.964}           & 0.983           \\

\bottomrule
\hline
\end{tabular}
\end{adjustbox}
\end{center}
\caption{Quantitative results on KITTI test data with Eigen split. Here, "Intr" indicates if camera intrinsics are learned or not during training, "-ref" meaning without refinement. The results are taken from their corresponding papers. Best results are highlighted in bold.}
\label{tab quant_res}
\end{table}

The KITTI benchmark has become a de facto standard for depth evaluation in recent times~\cite{guizilini20203d}. Hence, for evaluation, we make use of KITTI 2015 dataset~\cite{geiger2012we} which consists of 200 different scenes of driving data captured using RGB cameras along with sparse ground truth depth maps captured by a Velodyne LiDAR sensor. As the pre-processing step, we follow the previous works~\cite{zhou2017unsupervised} to remove static frames. This resulted in 39810 monocular raw images for training and 4424 for validation. In all the evaluation experiments, we make use of seven different metrics as proposed by Eigen \etal~\cite{eigen2014depth}, which are now commonly used and accepted evaluation indicators for comparison among various depth estimation approaches~\cite{zhao2020monocular}. A depth cap of 80 meters is used in all our experiments.

\textbf{Evaluation}. In this section, we compare the effectiveness of the proposed method with SOTA using KITTI test dataset and Eigen split~\cite{eigen2015predicting}. Eigen split contains 697 images with reprojected LIDAR points, which are used for evaluation. For depth evaluation, per-image median ground truth scaling (as proposed in~\cite{godard2019digging}) is utilized, to handle the unavailability of scale information in self-supervised monocular approaches~\cite{zhou2017unsupervised, godard2019digging}). No further post-processing steps were involved. We compare the results of our two variants, one without uncertainty estimation (model V1) and the other with uncertainty estimation (model V2) for evaluation on KITTI benchmark. As shown in \autoref{tab quant_res}, both our models achieved significant gains in both uncalibrated and calibrated scenarios in comparison to their corresponding baseline methods - OmegaNet~\cite{tosi2020distilled} and MonoDepth2~\cite{godard2019digging} respectively. In an uncalibrated scenario, where ground truth camera intrinsics are not known, our camera network predicts such values taking in a pair of images each time. Using such method, along with the proposed considerations, produced results which outperform the SOTA not only in uncalibrated but also in calibrated settings. In addition, a vivid improvement in depth map quality over baseline methods can be observed from qualitative results shown in \autoref{fig qual_res}. This shows the effectiveness of the proposed method, especially at object boundaries, primarily due to the proposed ESPCN layers. Notably, we achieved such competing metric scores using ResNext-50~\cite{xie2017aggregated} encoder, on contrary to high-end architectures used in~\cite{guizilini20203d} and~\cite{johnston2020self}.

%We compare the results of our two variants, one without uncertainty estimation (model V1) and the other with uncertainty estimation (model V2) with two divisions: one with learned intrinsics denoted by (L) and the other one without such learning (denoted by (N)).As shown in \autoref{tab quant_res}, both of our models are more accurate in comparison to the benchmark approaches: MonoDepth2~\cite{godard2019digging}, Omeganet~\cite{tosi2020distilled}, and other SOTA methods. Notably, our models manifest such results with only 25 Million(M) trainable parameters compared to other approaches like~\cite{guizilini20203d} and~\cite{johnston2020self} which use 128M and 88M parameters respectively. However, when compared in between, with Eigen split, version V1 is better in absolute relative error metric, version V2 in other metrics like RMSE which is sensitive to larger depth errors, and $\delta<1.25$, which represent near-range depth accuracy. Some samples of qualitative results comparing with our benchmarks can be seen in \autoref{fig qual_res}.

\begin{table}[!htbp]
\begin{center}
\begin{adjustbox}{max width=\textwidth}
\begin{tabular}{cccccccccc}
\hline
\toprule
& &\multicolumn{4}{c}{Lower is better} &\multicolumn{3}{c}{Higher is better} \\
\cmidrule(r){3-6}\cmidrule(l){7-9}
Method   & Resolution  &Abs Rel   &Sq Rel  &RMSE  &RMSE log  &$\delta$ $<$ 1.25  &  $\delta$ $<$ $1.25^{2}$  &$\delta$ $<$ $1.25^{3}$ \\
\bottomrule
\hline
 OmegaNet~\cite{tosi2020distilled}	& 640x192 &0.120	&0.792	&4.750	&0.191	&0.856	&0.958	&\textbf{0.984} \\
 MonoDepth2~\cite{godard2019digging} & 640x192   &0.115	&0.903	&4.863 	&0.193	&0.877	&0.959	&0.981  \\
 PackNet-SfM~\cite{guizilini20203d} &640x192  &0.111	&0.785	&4.601	&0.189	&0.878	&0.960	&0.982 \\
 HR Depth~\cite{lyu2020hr} & 640x192  &0.109	&0.792	&4.632	&0.185	&0.884	&0.962	&0.983 \\
 \textbf{Ours} & 640x192  &\textbf{0.106} 	&\textbf{0.750} 	&\textbf{4.482}	& \textbf{0.182}
 &\textbf{0.891}	&\textbf{0.964} 	&0.983  \\
 \midrule
 OmegaNet~\cite{tosi2020distilled}	 & 1024x320 &0.118	&0.748	&4.608	&0.186	&0.865	&0.961	&\textbf{0.985} \\
 MonoDepth2~\cite{godard2019digging}  & 1024x320 &0.115 &0.882 &4.701 &0.190 &0.879 &0.961 &0.982  \\
  PackNet-SfM~\cite{guizilini20203d} &1280 x 384  &0.107 &0.802 &4.538 &0.186 &0.889 &0.962 &0.981 \\
  HR Depth~\cite{lyu2020hr} & 1024x320   &0.106	&0.755	&4.472	&0.181	&0.892	&\textbf{0.966}	&0.984 \\
   HR Depth~\cite{lyu2020hr} &1280 x 384  &0.104 &0.727 &4.410 &0.179 &0.894 &\textbf{0.966} &0.984 \\

 \textbf{Ours} & 1024x320  &\textbf{0.102} &  \textbf{0.723} &  \textbf{4.374} &  \textbf{0.178} &  \textbf{0.898} &  \textbf{0.966} &  0.983 \\
\bottomrule
\hline
\end{tabular}
\end{adjustbox}
\end{center}
\caption{KITTI benchmark comparison with our baseline methods at different resolutions} %For 1024x320 resolution, only ResNet-18 encoder is used due to computational limitations. For 640x192 resolution, ResNext-50 depth encoder is used.}
\label{tab inp_res}
\end{table}

%\vspace{-5mm}

\begin{figure}[!t]
\begin{center}
\includegraphics[width=0.9\textwidth]{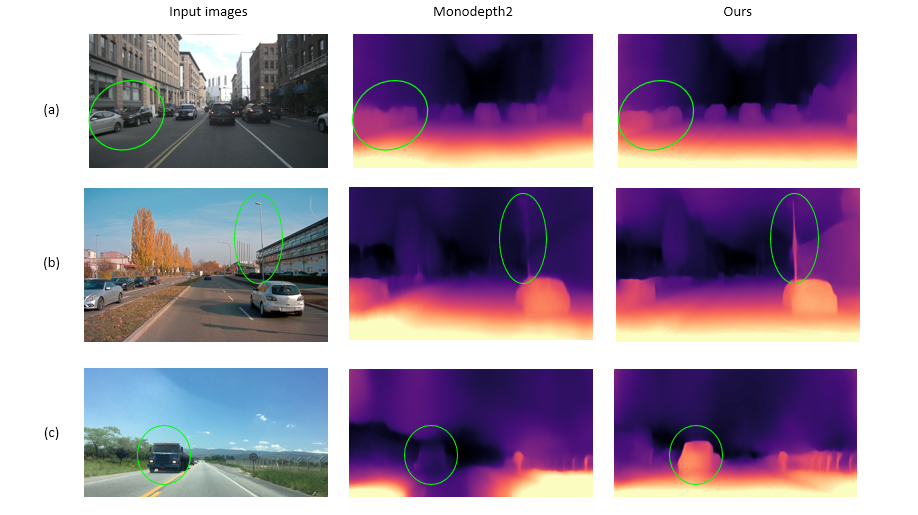}
\end{center}
\vspace{-15pt}
  \caption{Unseen images from (a) Nuscenes~\cite{caesar2020nuscenes}, (b) Audi a2d2~\cite{geyer2020a2d2} datasets respectively, and (c) a random video collected from \href{https://www.pexels.com/}{Pexels}. Better boundary separation (a), sharp results on thin objects (b) and accurate depth maps (c) suggest better applicability and generalizability of our model, trained just on KITTI dataset.}
\label{fig gen_cap}
\vspace{-15pt}
\end{figure}

\begin{table}[!htbp]
\begin{center}
\begin{adjustbox}{max width=\textwidth}
\begin{tabular}{cccccccccccc}
\hline
\toprule
& & & &\multicolumn{4}{c}{Lower is better} &\multicolumn{3}{c}{Higher is better} \\
\cmidrule(r){5-8}\cmidrule(l){9-11}
Method   & Intr  & SR & Uncert  &Abs Rel   &Sq Rel  &RMSE  &RMSE log  &$\delta$ $<$ 1.25  &  $\delta$ $<$ $1.25^{2}$  &$\delta$ $<$ $1.25^{3}$ \\
\bottomrule
\hline
 MD2 (R18)~\cite{godard2019digging}  & \xmark & \xmark  &\xmark    &0.115	&0.903	&4.863 	&0.193	&0.877	&0.959	&0.981 \\
 MD2 (R50)~\cite{godard2019digging}  & \xmark & \xmark  &\xmark    & 0.110 &0.831 &4.642 &0.187 &0.883 &0.962 &0.982    \\
 \midrule
  Ours  & \xmark & \xmark  &\xmark  &   0.111  &   0.761  &   4.743  &   0.188  &   0.876  &   0.960  &   0.983  \\
 Ours  & \cmark & \xmark  &\xmark  & 0.110 &  0.816 &  4.662 &  0.187 &  0.887 &  0.962 &  0.982 \\
 Ours  & \xmark & \cmark  &\xmark    &0.106 &  0.792 &  4.593 &  0.184 &  0.890 &  0.963 &  0.982 \\
 Ours  & \cmark & \cmark  &\xmark  &\textbf{0.105}	&0.768 	&4.539	&\textbf{0.182} 	&0.890	&\textbf{0.964}	&\textbf{0.983} \\
 Ours  & \cmark & \cmark  &\cmark   &0.106	&\textbf{0.750} 	&\textbf{4.482} 	&\textbf{0.182}	&\textbf{0.891}	&\textbf{0.964} 	&\textbf{0.983} \\
\bottomrule
\hline
\end{tabular}
\end{adjustbox}
\end{center}
\caption{Ablation studies on KITTI benchmark with baseline referring to~\cite{godard2019digging}, where “Intr” indicates training with intrinsics network. “SR” indicates the usage of super resolution, and “Uncert” stands for training with uncertainty.}
\label{tab abl.}
\vspace{-5mm}
\end{table}

\textbf{Input resolution}. In this experimental study, we have considered two common groups with image resolutions of 640x192 and 1024x320 pixels, respectively. In accordance to the reported results in \autoref{tab inp_res}, the proposed method (model V2) outperforms baselines (\cite{godard2019digging}, \cite{tosi2020distilled}) and other novel methods, at lower and higher resolutions. While using a higher input image resolution has not improved results significantly in the baseline methods, it has showed positive effect with our approach. Moreover, this improvement due to a change in resolution is highest among the group. This manifests the effectiveness of efficient sub-pixel convolutions at higher resolutions. Even when regressed to an inferior ResNet-18 pose encoder architecture at higher resolution owing to computational limitations, the results achieved along all performance metrics are the best among all our models.

\textbf{KITTI ablation experiments}. To evaluate the significance of each of the components proposed in our approach, we have performed ablation studies on KITTI test data with Eigen split (\autoref{tab abl.}). We evaluated depth by varying three main components in this study: (i) with intrinsics network implying that external feed of pre-calibrated camera intrinsic parameters is unnecessary, (ii) using efficient sub-pixel convolutions and, (iii) with uncertainty estimation. Our approach with the networks proposed and training camera intrinsics has itself improved results compared to baseline (here~\cite{godard2019digging}). It can be observed that our model trained without any of our components is not better than MonoDepth2, suggesting that these $3$ contributions lead to improvement. The main improvement is associated with the usage of efficient sub-pixel convolutions in the depth decoder, resulting in sharper depth maps at higher resolutions. Crucially, this improvement is even more valuable in terms of achieved performance when trained camera network. This finding is a key result for proving the effectiveness of our proposed method to leverage its usage for potentially unlimited datasets including videos from internet. Adding the heteroscedastic aleatoric uncertainty estimation, has not improved depth estimation significantly, where the results remained roughly the same.

\textbf{Generalization capability}. We perform qualitative evaluation using our model trained only on KITTI dataset, while testing on unseen data from unseen videos similar to~\cite{chen2019self}. However, we make use of two additional datasets - NuScenes~\cite{caesar2020nuscenes} and Audi a2d2~\cite{geyer2020a2d2}, along with a random road scene video from internet. Qualitative results of this experiment are demonstrated in \autoref{fig gen_cap}. Our model results are particularly exceptional along the object boundaries and in identifying thin objects compared to our baseline method~\cite{godard2019digging}. This evidence supports our claim that our method is generalizable to learn structures and scenes collected using different cameras, and also without the knowledge of camera intrinsics.

\vspace{-5pt}
\subsection{Videos from Internet}
\label{subsec:wild}
\begin{figure}[!htbp]
\vspace{-8pt}
\begin{center}
\includegraphics[width=\textwidth]{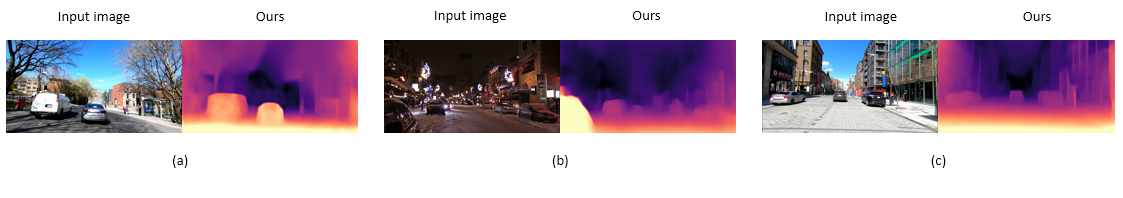}
\end{center}
\vspace{-30pt}
  \caption{Test results of our model, trained on multiple image sequences including KITTI and random internet videos.}
\label{fig wild_res}
\end{figure}

For demonstrating that our method can be used for videos even without ground-truth intrinsics, we have gathered 57 stock videos from the internet, predominantly covering road scenes which involve minimal distortion. The URLs of these videos will be made public. After pre-processing, it resulted in 7623 and 1373 images for training and test respectively. This dataset is particularly challenging as it involves different camera setups, also with varying camera heights, obstacles present throughout some videos (for~\eg car hood), highly varying image resolutions, and even weather conditions (day, rain, snow, night~\etc) as they are from the wild. %Sample images from this dataset are shown in~\autoref{fig wild_inp}%.
%These videos are collected randomly and for applications on real-time, they could be made more selective. 
For training, these images are resized to have a width of 640 and then, centre cropped to 640x192. This step assures that the input image is not stretched, although some vertical field of view is lost. Images from KITTI 2015 dataset~\cite{geiger2012we}, without intrinsics, involving similar pre-processing steps are used together for training along with this internet data to handle the variability and unavailability of large training dataset from the internet alone. Qualitative results, are demonstrated in \autoref{fig wild_res}, justify that the proposed approach can be used even when camera parameters are not known.

\vspace{-4mm}
\section{Conclusion}
\label{sec:concl}
\vspace{-2mm}
We propose a self-supervised monocular approach to learn depth for pinhole camera with minimum distortion, even without explicitly given intrinsics. This makes it possible to utilize videos from wild for training, which can be leveraged to obtain potentially unlimited data. Besides, we demonstrate that by incorporating ESPCN instead of interpolation in the depth decoder, robust and sharper depth maps can be obtained. In addition, we include an approach to estimate pixel-wise depth uncertainties, which could play a crucial role in robotics and autonomous driving tasks in tacking optimal decisions. Our contributions lead to significant improvement on KITTI benchmark even when ground truth camera intrinsics data is not available. Furthermore, through our experiments, we show that our best results are obtained at a higher input resolution of 1024x320 and to our knowledge, this result significantly outperforms the other state-of-the-art self-supervised monocular depth estimation methods on KITTI benchmark among methods which neither use ground truth data nor online refinement techniques for training.

\textbf{Acknowledgements}. This research has been supported by the computer vision team in project SpuMo E2D (funding reference number 03EFLSL015) under HTW Saar, and in part by the autonomous driving team (Agents and Simulated Reality department) of German Research Center for Artificial Intelligence (DFKI GmbH, Saarbrücken).

%-----------------------------------------------------------------------

\bibliography{main}
\end{document}

\maketitle
In this document, we provide additional evaluations and ablations experiments in support of our work in Section A and expand on the network architectures and implementation details in Section B.

\vspace{-10pt}
%-------------------------------------------------------------------------
\section{Additional Experiments}
\vspace{-5pt}
\subsection{Depth}

\textbf{KITTI Improved Ground Truth}. For Eigen split evaluation proposed by Eigen et al. \citep{eigen2014depth} on KITTI benchmark, the authors have made use of reprojected LIDAR points for depth evaluation. However, such approach does not account for other important aspects like occlusions, moving objects and ego-vehicle motion~\cite{godard2019digging}. Hence, we make use of improved ground truth from~\cite{uhrig2017sparsity}, in which occlusions are handled by considering stereo pairs and high quality depth maps which are produced by accumulating five consecutive frames. This results in 652 images, which accounts to 93$\%$ of images of the Eigen split. Depth is capped to 80 meters similar to Eigen split evaluation and the error metrics also remain the same. The evaluation results using such improved ground truth data are reported in \autoref{tab quant_res}, where both variants (V1 and V2) of our proposed method significantly outperform other benchmark methods in all evaluation metrics. Interestingly, when compared in between, version V2, namely the model with uncertainty estimation, exemplified better performance than model V1 without uncertainty estimation, in all the evaluation metrics. This infers the effectiveness of the uncertainty component on depth estimation.

\begin{table}[!htbp]
\begin{center}
\begin{adjustbox}{max width=0.9\textwidth}
\vspace{-15pt}
\begin{tabular}{ccccccccc}
\hline
\toprule
&\multicolumn{4}{c}{Lower is better}  &\multicolumn{3}{c}{Higher is better} \\
\cmidrule(r){2-5}\cmidrule(l){6-8}
Method     &Abs Rel     &Sq Rel    &RMSE   &RMSE log   &$\delta$ $<$ 1.25   &   $\delta$ $<$ $1.25^{2}$   &$\delta$ $<$ $1.25^{3}$ \\
\bottomrule
\hline
SFMlearner~\cite{zhou2017unsupervised} 	&0.176 &1.532 &6.129 &0.244 &0.758 &0.921 &0.971 \\
Vid2Depth~\cite{mahjourian2018unsupervised}	&0.134 &0.983 &5.501 &0.203 &0.827 &0.944 &0.981 \\
GeoNet~\cite{yin2018geonet}	&0.132 &0.994 &5.240 &0.193 &0.833 &0.953 &0.985 \\
DDVO~\cite{wang2018learning}   &0.126 &0.866 &4.932 &0.185 &0.851 &0.958 &0.986 \\
Ranjan~\cite{ranjan2019competitive} 	&0.123 &0.881 &4.834 &0.181 &0.860 &0.959 &0.985 \\
EPC++~\cite{luo2018every} 		&0.120 &0.789 &4.755 &0.177 &0.856 &0.961 &0.987 \\
MonoDepth2~\cite{godard2019digging}	&0.090 &0.545 &3.942 &0.137 &0.914 &0.983 &0.995 \\
\bottomrule
\hline
\textbf{Ours without uncert.(V1)}		&0.082  &   0.451  &   3.666  &   0.126  &   0.925  &   \textbf{0.986}  &   \textbf{0.996} \\ 
\textbf{Ours with uncert.(V2)}	&\textbf{0.081}  &   \textbf{0.427}  &   \textbf{3.532}  &   \textbf{0.124}  &   \textbf{0.928}  &   \textbf{0.986}  &   \textbf{0.996} \\ 
\bottomrule
\hline
\end{tabular}
\end{adjustbox}
\end{center}
\caption{The evaluation of results on KITTI test data with improved ground truth. The corresponding metrics are taken from their corresponding papers and~\cite{godard2019digging}, where the best results along each metric are highlighted in bold.}
\label{tab quant_res}
\end{table}

\begin{figure*}
\begin{center}
\includegraphics[width=6cm]{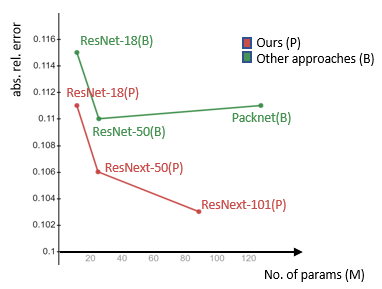}
\end{center}
   \caption{Comparison of baseline (B) methods with proposed (P) methods, with absolute relative error against number of trainable parameters in millions(M). Here ResNet-18, ResNet-50 versions of~\cite{godard2019digging} and Packnet from~\cite{guizilini20203d} are compared with ResNet-18, ResNext-50 and ResNext-101 depth encoders of our approach.}
\label{fig abs_params}
\end{figure*}

\textbf{Depth encoder variation}. Using high-end architectures for the depth estimation is advantageous, nevertheless, would come at the cost of higher training and inference times. Here, we report the impact of such architectures on depth performance by varying depth encoders. Therefore, we compare depth performance of ResNet-18(B), Resnet-50(B) architectures from~\cite{godard2019digging} and Packnet(B) architecture from,~\cite{guizilini20203d} with ResNet-18(P), ResNext-50 (P) and ResNext-101(P) architectures of our proposed method, where (B) indicates baseline approaches and (P) stands for our proposed method. All the ResNet and ResNext architectures (from both (B) and (P)) employ pretrained weights  on ImageNet~\cite{russakovsky2015imagenet} while Packnet~\cite{guizilini20203d} is not using any pretraining. From findings reported in \autoref{tab depth_enc}, the proposed method involving ResNet-18(P) depth encoder with just 11.69 million(M) trainable parameters, has achieved comparable performance in regard to the high-end baseline (B) methods including Packnet with 128 M trainable parameters. Meanwhile, the proposed approach with ResNext-50 (P) depth encoder outperforms all the baseline (B) approaches significantly, while ResNext-101(P) depth encoder improve results even further. Comparison of depth performance with various architectures and their corresponding trainable parameters is depicted in \autoref{fig abs_params}.

\begin{table}[!htbp]
\begin{center}
\begin{adjustbox}{max width=\textwidth}
\begin{tabular}{cccccccccc}
\hline
\toprule
& &\multicolumn{4}{c}{Lower is better}  &\multicolumn{3}{c}{Higher is better} \\
\cmidrule(r){3-6}\cmidrule(l){7-9}
Encoder     & param.(M)    &Abs Rel     &Sq Rel    &RMSE   &RMSE log   &$\delta$ $<$ 1.25   &   $\delta$ $<$ $1.25^{2}$   &$\delta$ $<$ $1.25^{3}$ \\
\bottomrule
\hline
 ResNet-18(B)  & 11.69  &0.115	&0.903	&4.863 	&0.193	&0.877	&0.959	&0.981   \\
 ResNet-50(B)  & 25.56     & 0.110 &0.831 &4.642 &0.187 &0.883 &0.962 &0.982    \\
 Packnet(B) 	&128	&0.111	&0.785	&4.601	&0.189	&0.878	&0.960	&0.982 \\
 \bottomrule
 \hline
 ResNet-18(P)     & 11.69  &   0.111  &   0.832  &   4.709  &   0.188  &   0.881  &   0.961  &   0.982    \\
 ResNext-50(P)       & 25.03  &0.106	&\textbf{0.750} 	&\textbf{4.482} 	&0.182	&0.891	&0.964 	&\textbf{0.983}    \\
 ResNext-101(P)     & 88.79  &\textbf{0.103}	&0.774	&4.514	&\textbf{0.180}	&\textbf{0.897}	&\textbf{0.965}	&\textbf{0.983} \\ 
\bottomrule
\hline
\end{tabular}
\end{adjustbox}
\end{center}
\caption{Table showing evaluation of results when the model is trained with different types of encoders. (B) indicates baseline and (P) indicates proposed method. Evaluation of results is done on KITTI test data. Except for the ResNet-18 approaches, for all other, ResNet-50 pose encoder is employed.}
\label{tab depth_enc}
\end{table}

\begin{figure}[!htbp]
\begin{center}
\includegraphics[width=\textwidth]{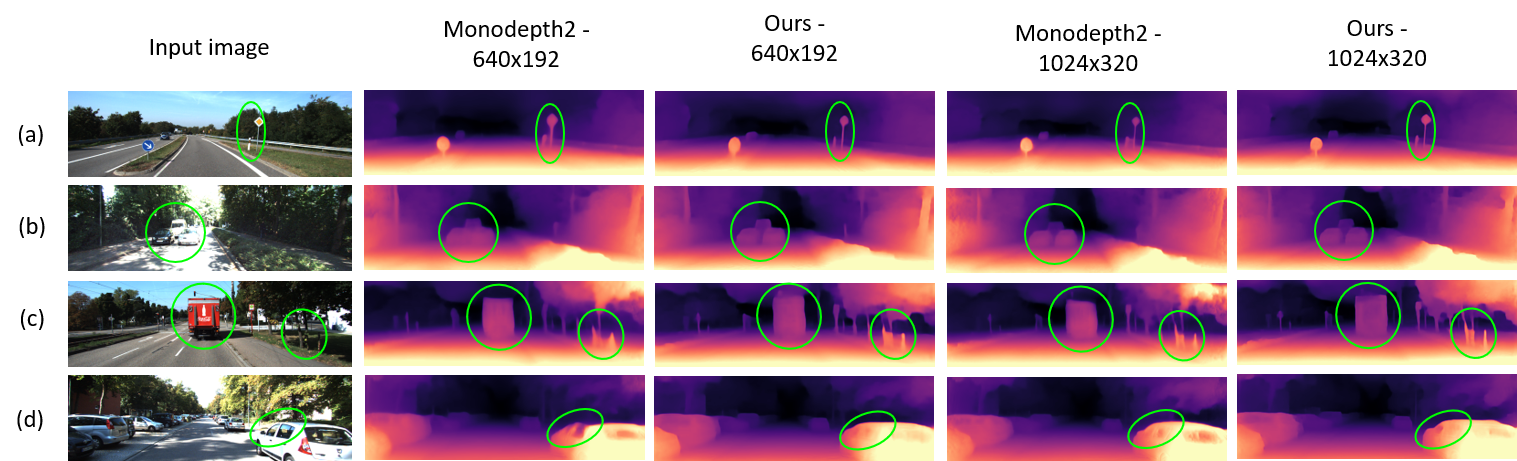}
\end{center}
  \caption{The qualitative comparison of our approach in contrary to MonoDepth2~\cite{godard2019digging} with (a) near thin objects, (b) with close objects, (c) at object boundaries, (d) colour saturated regions.}
\label{fig comp_mono}
\end{figure}

\begin{figure}[!htbp]

\begin{center}
\includegraphics[width=0.8\textwidth]{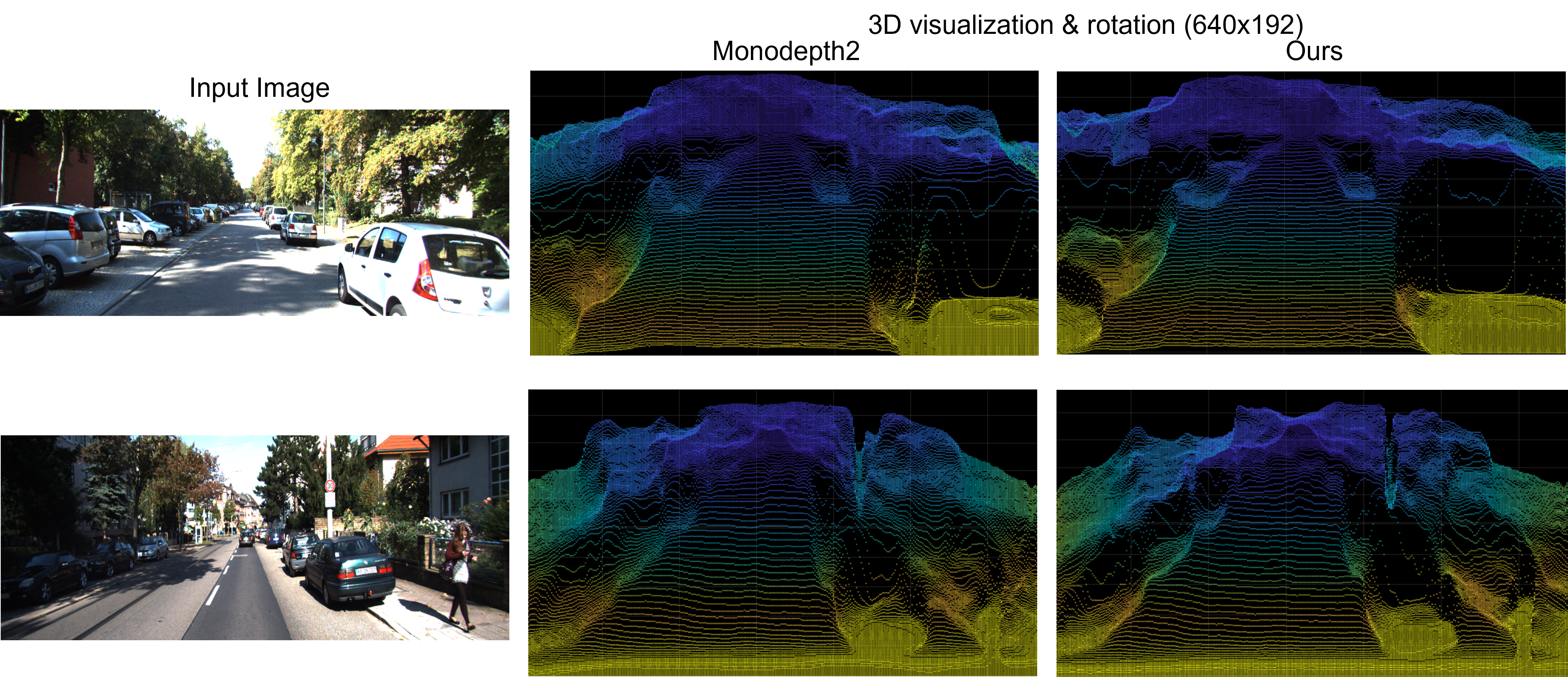}
\end{center}
  \caption{The 3D qualitative comparison of our approach in contrary to MonoDepth2~\cite{godard2019digging} in 3D (X, Y, Z as in point cloud datasets) format, where the normalized color value for every single pixel from the predicted images is used as depth (Z) value during 3D visualization. In addition, the rotation of the point cloud data has been applied during renderings in order to emphasize the improvements of the proposed approach over the baseline, namely less uncertainty on neighbour object boundary regions as well as more accurate reconstruction of shapes.}
\label{fig comp_mono_3d}
\end{figure}

\textbf{Comparison with MonoDepth2}~\cite{godard2019digging}. In this experiment, we compare our model results with our baseline, MonoDepth2 ~\cite{godard2019digging} to show the significance of our approach. \autoref{tab comp_mono} shows the results of baseline and the proposed method at both, lower (640x192) and higher input resolutions (1024x320) images. Here, the baseline~\cite{godard2019digging} has not shown greater improvement with change in input resolution, nevertheless, our approach manifested significant improvement due to our use of sub-pixel convolutions for upsampling. We have also compared our approach with baseline at edge cases involving thin objects, object boundaries and colour saturated regions, and this qualitative analysis is demonstrated in \autoref{fig comp_mono}. Our model, especially for higher resolution input, exemplified better depth results at thin objects and the objects nearer to the camera. Also, sharper depth results can be observed at object boundaries, credit to our efficient sub-pixel convolutions which enabled better super-resolution for upsampling in between decoder layers. In colour saturated regions (such as in \autoref{fig comp_mono}(d)), the low resolution model of MonoDepth2 shows notable artifacts which are somewhat compensated with the higher resolution model. Both our models, however, do not exhibit any such artifacts and result in smoother and robust depth maps compared to the baseline approach. To support the above listed statements, we additionally rendered 3D (X,Y,Z) perspectives for the generated depth images (\autoref{fig comp_mono_3d}), where Z coordinate stands for the normalized value for each pixel of the generated depth image. As seen from \autoref{fig comp_mono_3d} our approach shows less uncertainty (less dense points) in boundary regions of the objects. In other words, the proposed approach predicts better the class assignment for each depth point, e.g. vehicle vs background. Moreover, it demonstrates a better accuracy in terms of the reconstructed shape of scene objects.

\begin{table}[!htbp]
\begin{center}
\begin{adjustbox}{max width=\textwidth}
\begin{tabular}{cccccccccc}
\hline
\toprule
& &\multicolumn{4}{c}{Lower is better}  &\multicolumn{3}{c}{Higher is better} \\
\cmidrule(r){3-6}\cmidrule(l){7-9}
Method     & Resolution    &Abs Rel     &Sq Rel    &RMSE   &RMSE log   &$\delta$ $<$ 1.25   &   $\delta$ $<$ $1.25^{2}$   &$\delta$ $<$ $1.25^{3}$ \\
\bottomrule
\hline
MonoDepth2~\cite{godard2019digging}  & 640x192     &0.115	&0.903	&4.863 	&0.193	&0.877	&0.959	&0.981   \\
Ours (V2)	& 640x192	&\textbf{0.106}	&\textbf{0.750} 	&\textbf{4.482} 	&\textbf{0.182}	&\textbf{0.891}	&\textbf{0.964} 	&\textbf{0.983} \\
\hline
MonoDepth2~\cite{godard2019digging} & 1024x320     &0.115 &0.882 &4.701 &0.190 &0.879 &0.961 &0.982    \\
Ours (V2) & 1024x320  &\textbf{0.102}  &   \textbf{0.723}  &   \textbf{4.374}  &   \textbf{0.178}  &   \textbf{0.898}  &   \textbf{0.966}  &   \textbf{0.983}  \\
\bottomrule
\hline
\end{tabular}
\end{adjustbox}
\end{center}
\caption{The evaluation results on KITTI benchmark with change in input resolution. For 1024x320 resolution, only ResNet-18 pose encoder is used due to computational limitations. For 640x192 resolution, ResNext-50 pose encoder is used.}
\label{tab comp_mono}
\end{table}

\vspace{-10pt}
\subsection{Odometry Evaluation}
\label{subsec:odom_eval}
\textbf{KITTI Odom Dataset}. KITTI Odometry dataset~\cite{Gomez2015} consists of 22 driving sequences, of which only 11 sequences have ground truth trajectories and the remaining 11 sequences are unlabelled. We make use of only labelled sequences for both, training and evaluation. Similar to the prior approaches~\cite{godard2019digging, zhou2017unsupervised}, we first train our models on sequences - 00 to 08 using the KITTI Odometry split which has 36630 images and then evaluate separately on sequence 09 (1591 images) and sequence 10 (1201 images) data. 

\textbf{Performance Metrics}. For odometry evaluation, we use Absolute Trajectory Error (ATE) as performance metric as proposed in~\cite{zhou2017unsupervised}. This metric computes the Root Mean Square Error (RMSE) between the ground truth and the estimated trajectory. ATE can be computed with any of these: translation, rotation or velocity. All these parameters return the same single error metric, making it easier to compare~\cite{zhang2018tutorial}. We opt translation to compute ATE similar to previous approaches~\cite{godard2019digging} for odometry evaluation.

\textbf{Evaluation}. For odometry evaluation, we follow~\cite{godard2019digging} and predict on five-frame test sequence, which is used in~\cite{zhou2017unsupervised}. We calculate the ATE of our predictions for each of the four pairs in a five-frame test sequence, and then combine them to report the metric's mean and standard deviation. Results of odometry evaluation are reported in \autoref{tab odom_eval}. For this experiment, we have employed default settings of the proposed method. Our proposed approach exemplifies similar performance with our baseline~\cite{godard2019digging}. There is a significant improvement in depth results when compared with our baseline, however, not in pose, which is predominantly impacted by the use of common encoder for both camera pose and intrinsics estimation. In addition, when compared with other approaches~\cite{mahjourian2018unsupervised, yin2018geonet, ranjan2019competitive}, similar to~\cite{godard2019digging}, the improvement is not significant because of (i) not using custom architectures for odometry evaluation as this particular pose network is designed particularly for better depth estimation and (ii) using only two frames for pose estimation.

\begin{table}[!htbp]
\begin{center}
\begin{adjustbox}{max width=\textwidth}
\begin{tabular}{cccc}
\hline
\toprule
Method &     $\#$frames & Sequence 09  &  Sequence 10     \\
\bottomrule
\hline
DDVO \citep{wang2018learning} &3 &0.045±0.108 &0.033±0.074 \\ 
SFMlearner \citep{zhou2017unsupervised} &5 &0.021±0.017 &0.020±0.015 \\
Mahjourian \citep{mahjourian2018unsupervised} &3 &0.013±0.010 &0.012±0.011 \\
GeoNet \citep{yin2018geonet} &3 &\textbf{0.012±0.007} &0.012±0.009 \\
EPC++ \citep{luo2018every} &3 &0.013±0.007 &\textbf{0.012±0.008} \\
Ranjan \citep{ranjan2019competitive} &3 &\textbf{0.012±0.007} &\textbf{0.012±0.008} \\
MonoDepth2 \citep{godard2019digging} &2 &0.017±0.008 &0.015±0.010 \\
\textbf{Ours}       &2    &0.018±0.009   & 0.015±0.009  \\
\bottomrule
\hline
\end{tabular}
\end{adjustbox}
\end{center}
\caption[Pose evaluation]{Pose evaluation performed with KITTI Odom data. Sequences 09 and 10 of this dataset are used for evaluation purposes, where $\#$frames indicate the number of input frames for pose network. The mean ATE along with its standard deviation is reported for each approach.}
\label{tab odom_eval}
\end{table}

\vspace{-15pt}
\subsection{Camera Intrinsics Evaluation}

We opt for KITTI odometry dataset~\cite{Gomez2015} for camera intrinsics evaluation similar to our odometry evaluation (refer \autoref{subsec:odom_eval}). Here, the model is trained with sequences - 00 to 08, and we use sequences 09 and 10 for evaluation. The mean and standard deviation of focal length parameters ($f_{x}$ and $f_{y}$) and principal offsets ($x_{0}$ and $y_{0}$) are reported with each of the evaluation sequences, and these parameters are compared with the ground truth data. Distortion parameters are not modelled, hence are not reported. For this setup, we use ResNext-50 depth encoder and ResNet-18 pose encoder, with input resolution to the framework as 1024x320. The quantitative evaluation findings for this experiment are shown in \autoref{tab intr_eval}. It can be observed that the results are impacted significantly by using a light-weight camera network and also due to the  usage of a common encoder for camera pose and intrinsics. Nevertheless, the depth estimation results were improved, with this intrinsics estimations, when compared to given camera intrinsics as input (shown in KITTI ablation experiments as part of the main paper).

\begin{table}[!htbp]
\begin{center}
\begin{adjustbox}{max width=\textwidth}
\begin{tabular}{cccc}
\hline
\toprule
Method &   Sequence 09  &   Sequence 10   & Ground Truth   \\
\bottomrule
\hline
Horizontal focal length ($f_{x}$)   & 0.6902 ± 0.0138     &0.6868 ± 0.0208      &0.5767    \\
Vertical focal length ($f_{y}$)     & 1.9208 ± 0.0805      &1.8947 ± 0.0807      &1.9111      \\
Horizontal centre ($x_{0}$)         & 0.5032 ± 0.0031     &0.5017 ± 0.0029       &0.4909      \\
Vertical centre ($y_{0}$)           & 0.4681 ± 0.0025      &0.4678 ± 0.0030       &0.4949     \\
\bottomrule
\hline
\end{tabular}
\end{adjustbox}
\end{center}
\caption{Our camera network estimates of the intrinsics parameters, normalized by the input image resolution, are compared against the ground truth data. The mean of these normalized intrinsics estimates with standard deviations are reported for each parameter computed on all the images from KITTI Odom 09 and 10 sequences.}
\label{tab intr_eval}
\vspace{-15pt}
\end{table}

\vspace{-8pt}
\section{Additional Implementation Details}

\begin{figure}[!t]
\begin{center}
%\vspace{-15pt}
\includegraphics[width=0.95\textwidth]{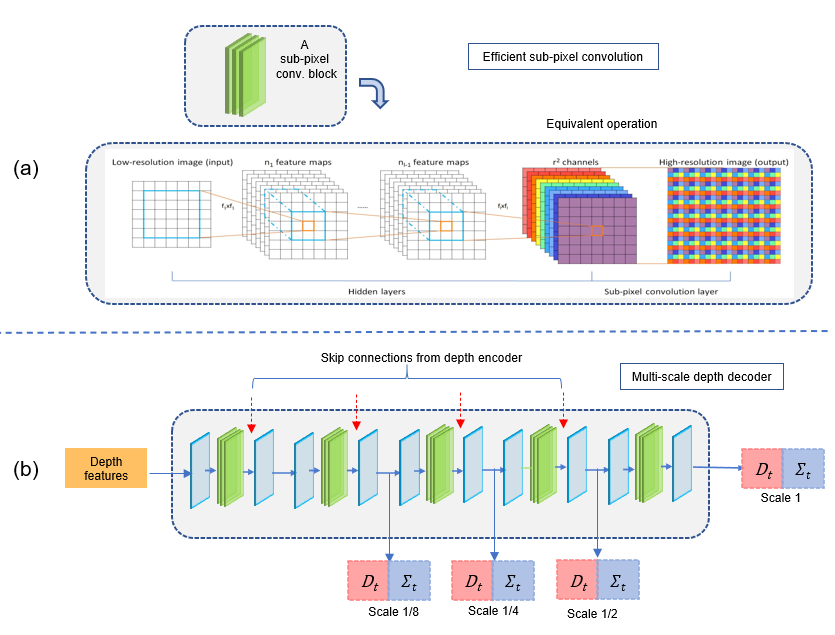}
\end{center}
  \caption{(a) shows the sub-pixel convolution operation adapted from~\cite{shi2016real} depicting three 3x3 2D convolutions followed by a pixel-shuffling operation which rearranges pixels from higher number of channels into higher resolution of width and height. (b) shows the multiscale decoder architecture with 3x3 convolutions (blue) and sub-pixel convolutions (green) predicting outputs at four scales. }
\label{fig dec_n_sub_pix}
\vspace{-5pt}
\end{figure}

\begin{table}[!htbp]
\begin{center}
\begin{adjustbox}{max width=\textwidth}
\begin{tabular}{cccccccc}
\hline
\toprule
layer     &k       &s      &p      &chns   &res    &input    &activation   \\
\bottomrule
\hline
dconv5     &3     &1     &1     &256     &32      &econv5      &ELU   \\
s1conv5     &5     &1     &2     &64     &32      &dconv5      &ReLU   \\
s2conv5    &3      &1     &1     &32     &32      &s1conv5      &ReLU  \\
upconv5    &3      &1     &1     &256*4     &32      &s2conv5      &ReLU   \\
iconv5     &3     &1     &1      &256     &16      & [ps]upconv5, econv4      &ELU   \\
\midrule
dconv4     &3     &1     &1     &128     &16      &iconv5      &ELU   \\
s1conv4     &5    &1     &2      &64     &16      &dconv4      &ReLU   \\
s2conv4    &3     &1     &1      &32     &16      &s1conv4      &ReLU  \\
upconv4     &3    &1     &1      &128*4     &16      &s2conv4      &ReLU   \\
iconv4     &3     &1     &1     &128     &8      &[ps]upconv4, econv3     &ELU   \\
disp\_uncert4     &3      &1     &1     &2     &1      & iconv4      &Sigmoid   \\
\midrule
dconv3     &3     &1     &1     &64     &8      &iconv4      &ELU   \\
s1conv3     &5    &1     &2      &64     &8      &dconv3      &ReLU   \\
s2conv3     &3    &1     &1      &32     &8      &s1conv5      &ReLU  \\
upconv3     &3    &1     &1      &64*4     &8      &s2conv5      &ReLU   \\
iconv3     &3     &1     &1     &64     &4      &[ps]upconv3, econv2      &ELU   \\
disp\_uncert3     &3      &1     &1     &2     &1      &iconv3      &Sigmoid   \\
\midrule
dconv2     &3     &1     &1     &32     &4      &iconv3      &ELU   \\
s1conv2     &5    &1     &2      &64     &4      &dconv2      &ReLU   \\
s2conv2     &3    &1     &1      &32     &4      &s1conv5      &ReLU  \\
upconv2     &3    &1     &1      &32*4     &4      &s2conv5      &ReLU   \\
iconv2     &3     &1     &1     &32     &2      &[ps]upconv2, econv1      &ELU   \\
disp\_uncert2     &3     &1     &1     &2     &1      &iconv2      &Sigmoid   \\
\midrule
dconv1     &3    &1     &1      &16     &2      &iconv2      &ELU   \\
s1conv1     &5    &1     &2      &64     &2      &dconv1      &ReLU   \\
s2conv1    &3     &1     &1      &32     &2      &s1conv1      &ReLU  \\
upconv1     &3    &1     &1      &16*4     &2      &s2conv1      &ReLU   \\
iconv1     &3     &1     &1     &16     &1      &[ps]upconv1     &ELU   \\
disp\_uncert1     &3     &1      &1     &2     &1      &iconv1      &Sigmoid   \\
\bottomrule
\end{tabular}
\end{adjustbox}
\end{center}
\caption{The network details of depth decoder used in our approach. Here, $k$ indicates kernel size, $p$ indicates padding, $chns$ indicate number of channels for that layer, $res$ stands for the downsampling factor, where $1$ indicates full resolution, $input$ stands for the input to that layer, [ps] indicates pixel shuffle operation with an upsample factor of 2, $econv$ represents inputs from various levels of the encoder, and lastly $activation$ indicates the activation function used for that corresponding layer.}
\label{tab depth_net}
\vspace{-7pt}
\end{table}

\textbf{Depth Network}. We embed efficient sub-pixel convolutions into the depth decoder for better upsampling in contrary to the nearest-neighbour interpolation followed in previous approaches~\cite{godard2019digging, zhou2017unsupervised, wang2018learning, yin2018geonet}. These efficient sub-pixel convolutions involve three convolution operations followed by a final pixel shuffling operation, which operates at lower resolution in extracting feature information necessary to perform super-resolution. Just before the final pixel shuffle operation, the output is arranged to have the channels multiplied by a factor of, $r^2$ where $r$ is the upsample factor. This output is finally shuffled across the pixels along channels to obtain image super-resolution.This operation is shown in \autoref{fig dec_n_sub_pix}(a).

The depth network, in overall, takes in a single image as input and outputs disparity or inverse-depth along with a pixel-wise uncertainty map. These outputs are obtained at four scales, \ie at 1/8, 1/4, 1/2 and 1, scaling with input image resolution as shown in \autoref{fig dec_n_sub_pix}(b). Such multi-scale decoder is employed to prevent the gradient locality of the bilinear sampler and to prevent the loss objective getting stuck at local minimum~\cite{godard2019digging}. The disparity map obtained is later converted to depth using $D = 1/(a*\sigma+b)$ where $D$ represents depth and $\sigma$ represents the disparity map. Here, $a$ and $b$ are constants which are chosen such as to constrain the depth values between 0.1 and 100. The detailed network architecture is shown in \autoref{tab depth_net}.

\begin{table}[!htbp]
\begin{center}
\begin{adjustbox}{max width=\textwidth}
\begin{tabular}{cccccccc}
\hline
\toprule
\multicolumn{8}{c}{Pose Decoder} \\
\bottomrule
\hline
layer     & k       & s      & p      & chns   & res    & input    & activation   \\
\bottomrule 
\hline
pconv0     &1     &1     &1     &256     &32      &econv5      &ReLU   \\
pconv1     &3     &1     &1     &256     &32      &pconv0      &ReLU   \\
pconv2    &3      &1     &1     &256     &32      &pconv1      &ReLU  \\
pconv3    &1      &1     &1     &6     &32      &pconv2      &-   \\
\hline
\toprule
\multicolumn{8}{c}{Camera Network} \\
\bottomrule
\hline
inconv0     &1     &1     &1     &256     &32      &econv5      & ReLU   \\
inconv1     &3     &1     &1     &2     &32      &inconv0      & Softplus   \\
inconv2    &3      &1     &1     &2     &32      &inconv0      &-  \\
inconcat    &-      &-     &-     &-     &-     &inconv1, inconv2       &-   \\
\bottomrule
\hline
\end{tabular}
\end{adjustbox}
\end{center}
\caption[Network details of the pose decoder and the camera network]{The network details of the pose decoder and the camera network are shown. Here, $k$ indicates the kernel size, $p$ indicates the padding parameter, $chns$ stands for the number of channels for that particular convolutional layer, $res$ corresponds to the downsampling factor, where 1 indicates full resolution, $input$ stands for the input to that layer, $econv5$ is the output from pose encoder, and lastly $activation$ shows the used activation function}
\label{tab pose_n_cam}
\vspace{-8pt}
\end{table}

\textbf{Pose and Camera Networks}. The pose encoder is modified to take in a pair of images as input, which are concatenated channel-wise. Also, the weights in the expanded filter are divided by 2, which makes the output of the convolution similar to ResNet default output with a single image input~\cite{godard2019digging}. The camera network follows a light-weight architecture to estimate camera intrinsic parameters. Horizontal focal length ($f_{x}$) and Horizontal centre ($x_{0}$) parameters are initialized to $W/2$, and Vertical focal length ($f_{y}$) and Vertical centre ($y_{0}$) parameters are initialized to $H/2$ for better convergence following~\cite{gordon2019depth}, where $W$ and $H$ represents width and height of the input image respectively. Network details of both pose and camera networks are shown in \autoref{tab pose_n_cam}.
\vspace{-5pt}
%-------------------------------------------------------------------------

\bibliography{main}